\documentclass[conference]{IEEEtran}
\IEEEoverridecommandlockouts
\usepackage{mathtools}
\usepackage{amsmath,amssymb,amsfonts}
\usepackage{algorithm}
\usepackage{amsthm}
\usepackage{cancel}
\usepackage{algorithmic}
\usepackage{hyperref}
\usepackage[square,sort,comma,numbers]{natbib}
\DeclareMathOperator*{\argmax}{arg\,max}

\usepackage{comment}
\usepackage{dsfont}

\usepackage{graphicx}
\usepackage{textcomp}
\newtheorem{theorem}{Theorem}
\newtheorem{lemma}[theorem]{Lemma}
\usepackage{xcolor}
\def\BibTeX{{\rm B\kern-.05em{\sc i\kern-.025em b}\kern-.08em
    T\kern-.1667em\lower.7ex\hbox{E}\kern-.125emX}}
\begin{document}

\title{Online Reinforcement Learning in Periodic MDP\\ 
\thanks{The authors are with the Department of Electrical Engineering
IIT Delhi, . Email: \{Ayush.Aniket, arpanc\}@ee.iitd.ac.in . }
}

\author{\IEEEauthorblockN{ Ayush Aniket}
\and
\IEEEauthorblockN{ Arpan Chattopadhyay}
}

\maketitle

\begin{abstract}
We study learning in periodic Markov Decision Process (MDP), a special type of non-stationary MDP where both the state transition probabilities and reward functions vary periodically, under the average reward maximization setting. We formulate the problem as a stationary MDP by augmenting the state space with the period index, and propose  a periodic upper confidence bound  reinforcement learning-2 (PUCRL2) algorithm. We show that the  regret of PUCRL2 varies linearly with the period $N$ and as $\mathcal{O}(\sqrt{Tlog T})$ with the horizon length $T$. Utilizing the information about the sparsity of transition matrix of augmented MDP, we propose another algorithm PUCRLB which enhances upon PUCRL2, both in terms of regret  ($O(\sqrt{N})$ dependency on period) and empirical performance. Finally, we propose two other algorithms  U-PUCRL2 and U-PUCRLB for extended uncertainty in the environment in which the period is unknown but a set of candidate periods are known. Numerical results demonstrate the efficacy of all the algorithms.
\end{abstract}


\section{Introduction}
Reinforcement learning (RL) deals with the problem of optimal sequential decision making in an unknown environment. Sequential decision making in an environment with an unknown statistical model  is typically modeled as a Markov decision process (MDP) where the decision maker, at each time step $t$, has to take an action $a_t$ based on the state $s_t$ of the environment, resulting in a probabilistic  transition to the next state $s_{t+1}$ and a reward $r_t$ accrued by the decision maker depending on the current state and current action.  RL has applications in many areas including robotics \cite{kober2013reinforcement}, resource allocation in wireless networks \cite{5137416}, finance \cite{bacoyannis2018idiosyncrasies} etc.

In a stationary MDP, the unknown transition probabilities and reward functions are invariant with time. However, the ubiquitous presence of non-stationarity in real world scenarios often limits the application of stationary reinforcement learning algorithms. Most of the existing works require information about the maximum possible amount of changes that occur in the environment via variation budget in the transition and reward function, or via the number of times the environment changes; this does not require any assumption on the nature of non-stationarity in the environment. On the contrary, we consider a periodic MDP whose state transition probabilities and reward functions are unknown but periodic with a known period $N$. In this setting, we propose PUCRL2 and PUCRLB algorithms and analyse their regret. Also, for a setting in which the period is unknown, we propose two other  algorithms U-PUCRL2 and U-PUCRLB and demonstrate their performance via simulation.

Non-stationary RL has been extensively studied in a variety of scenarios \cite{auer2008near,gajane2018sliding,li2019online,ortner2020variational,cheung2020reinforcement,fei2020dynamic,domingues2021kernel,mao2021near,zhou2020nonstationary,touati2020efficient,wei2021non}.The authors of \cite{auer2008near} propose a restart version of the popular UCRL2 algorithm  meant for stationary RL problems, which achieves an $\mathcal{\Tilde{O}}(l^{1/3}T^{2/3})$ regret where $T$ is the number of time steps, under the setting in which the MDP changes at most $l$ number of times. In the same setting \cite{gajane2018sliding} shows that UCRL2 with sliding windows achieves the same regret. In time-varying environment, a more apposite measure for performance of an algorithm is dynamic regret which measures the difference between accumulated reward through online policy and that of the optimal offline non-stationary  policy. This was first analysed in \cite{li2019online} in a solely  reward varying environment. The authors of   \cite{ortner2020variational} propose first variational dynamic regret bound of $\mathcal{\Tilde{O}}(V^{1/3}T^{2/3})$, where $V$ represents the total variation in the MDP. The work of \cite{cheung2020reinforcement} provides the sliding-window UCRL2 with confidence widening, which achieves an
$\mathcal{\Tilde{O}}((B_r+B_p)^{1/4}T^{3/4})$ dynamic regret, where $B_r$ and $B_p$ represent the maximum amount of possible variation in reward function and transition kernel respectively. They also propose a Bandit-over-RL (BORL) algorithm which tunes the UCRL2-based algorithm in the setting of unknown variational budgets. Further, in the model-free and episodic setting,  \cite{wei2021non} propose  policy optimization algorithms and \cite{fei2020dynamic} propose RestartQ-UCB which achieves a dynamic regret bound  of $\mathcal{\Tilde{O}}(\Delta^{1/3}HT^{2/3})$,where $\Delta$ represent the amount of changes in the MDP and H represents the episode length. The paper \cite{domingues2021kernel} studies a kernel based approach for non-stationarity in MDPs with metric spaces. In the linear MDP case, \cite{mao2021near} and \cite{zhou2020nonstationary} provide optimal regret guarantees.
 Finally the authors of \cite{wei2021non} provide a black-box algorithm which turns any (near)-stationary algorithm to work in a non-stationary environment with optimal dynamic regret  $\Tilde{O}(\min{\sqrt{LT},\Delta^{1/3}T^{2/3}})$, where $L$ and $\Delta$ represent the number and amount of changes of the environment,  respectively.
 
 Periodic MDP (PMDP)  has been marginally studied in literature.  The authors of \cite{riis1965discounted}   study it in the discounted reward setting, where a policy-iteration algorithm is proposed. The authors of   \cite{veugen1983numerical} propose the first state-augmentation method for conversion of PMDP into a stationary one, and  analyse the performance of various iterative methods for finding the optimal policy. Recently, \cite{hu2014near} derive a corresponding value iteration algorithm suitable for periodic problems in discounted reward case and provide near-optimal bounds for greedy periodic policies. To  our knowledge, RL in  PMDP has not been studied. 
 
 In this paper, we make the following contributions: 
 \begin{itemize}
     \item  In Section~\ref{section:algorithms}, we study a special form of non-stationarity where the unknown reward and transition functions vary periodically with a known period $N$. We propose a modification PUCRL2 of UCRL2, which treats the periodic MDP as stationary MDP with augmented state space. We derive a static regret bound which has  a linear dependence on $N$ and sub-linear dependence on $T$. 
    \item By utilizing the information about the sparsity of the transition matrix of augmented MDP, we propose another algorithm PUCRLB, a variant of UCRLB.  PUCRLB achieves a better regret bound than PUCRL2; its regret has a $\sqrt{N}$ dependence on period, ( Section~\ref{section:algorithms}).
    \item Further, in Section~\ref{section:unknown-N}, we study an extended uncertainty environment wherein the period information is unknown and hidden among a set of  candidate periods. We propose two algorithms U-PUCRL2 and U-PUCRLB, and demonstrate their performance numerically in Section~\ref{section:numerical}.

\end{itemize}

\section{Problem Formulation}
\label{Problem Formulation}

We consider a discrete time PMDP with a finite state space $\mathcal{S}$ where $\lvert \mathcal{S} \rvert = S$, a finite action space $\mathcal{A}$ where $\lvert \mathcal{A} \rvert = A$. $N \geq 2$ is an integer value representing the period of the PMDP. $p_i(s'|s,a)\hspace{+1mm} \forall (s',s,a) \in \mathcal{S}\times\mathcal{A} \times \mathcal{S}$ is the probability for the next state given current state-action pair,  and $r_i(s,a)\hspace{+1mm} \forall (s,a) \in \mathcal{S}\times\mathcal{A}$ is the mean reward given current state-action pair, for all period indices $ i \in \{1,2,..,N\}$.

Let us define $\mathbf{P}_{t}(s,a)$ as the transition probability matrix for a given $(s,a)$ pair at time $t$. By the periodicity assumption, $\mathbf{P}_{t+N}(s,a) = \mathbf{P}_t(s,a) $ and $r_{t+N}(s,a) = r_t(s,a) \hspace{+1mm} \forall (s,a) \in \mathcal{S}\times\mathcal{A} \hspace{+1mm}, \forall t\geq 1 $.  The time horizon length is   $T >> N$.

Now, the PMDP can be transformed into a stationary MDP with augmented state-space (henceforth referred as AMDP). In this AMDP, we couple the period index and states together to obtain an augmented state space $\mathcal{S'} =  \mathcal{S} \times \{1,2,...N\}$; if the state of the original MDP is $s$ at time $t$, then the corresponding state in the AMDP will be $(s, ((t-1)  \mod{N})+1)$, where $\mod$ represents the modulo operator. Consequently, the (time-homogeneous) transition probability of the AMDP for current state $s$ and current action $a$ becomes:
\begin{equation*}
p((s',n') | (s,n),a) = \left\{
        \begin{array}{ll}
            0 & n' \neq n+1 \mod{N}\\
            p_n(s'|s,a) & n' = n+1 \mod{N} \\
        \end{array}
    \right.
\end{equation*}

\begin{figure} 
  \centering
  \includegraphics[height=5cm, width=6cm]{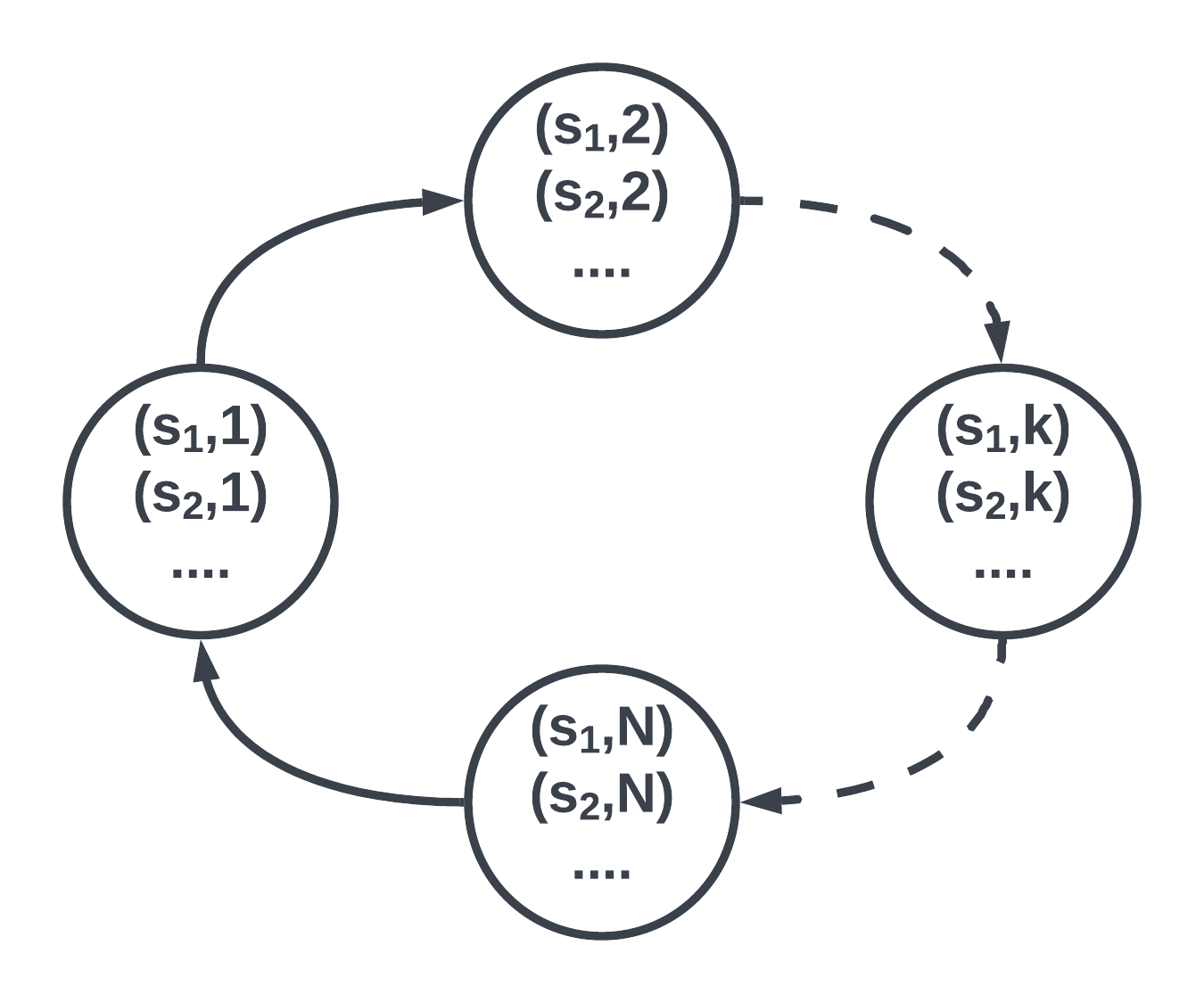}
  \vspace{-6mm}
  \caption{Augmented MDP with periodic states.}
  \label{fig:AMDP}
  \vspace{-5mm}
\end{figure}

The corresponding mean reward of the AMDP is given by  $r((s,n),a) = r_{n}(s,a)$. The probability mass function of the next augmented state given current (state, period)-action pair is denoted by $\mathbf{p}(\cdot | (s,n),a)$. Obviously, under any deterministic stationary policy for the AMDP,  each (state, period index) pair can only be visited after $N$ number of time steps. Thus, the PMDP becomes a stationary AMDP with periodic transition matrix as shown in Figure \ref{fig:AMDP}. Let $\rho^*$ denote the optimal time-averaged (average expected reward over large number of time steps and then taking a Cesaro limit) reward \cite[Section 8.2.1]{puterman2014markov} of the AMDP. In this paper, we seek to develop an RL algorithm so as to minimize the static regret with respect to this optimal average reward $\rho^*$. Let $\pi$ be any generic  policy for the AMDP. Our problem is to minimize the expected {\em static} regret over all policies:
  \begin{equation*} \label{eqn:regret}
  \min_{\pi}  \sum_{t =1}^{T} (\rho^* - \mathbb{E}_{\pi} (r_t((s_t,n_t),a_t)))
 \end{equation*}

\section{Algorithms for known period}
\label{section:algorithms}
In this section, we propose two algorithms named PUCRL2 and PUCRLB for PMDP with known $N$.  While PUCRL2 is motivated by UCRL2 algorithm, the PUCRLB algorithm is  developed to handle the sparsity coming from the state augmentation operation.

\subsection{PUCRL2 algorithm}
PUCRL2 (Algorithm~\ref{alg:PUCRL2}) estimates the mean reward and the transition kernel for each augmented state-action pair, while keeping in mind that the transition occurs only to augmented states with the next period index and the probability of transitioning to other augmented states is zero. Hence, the algorithm only estimates the non-zero transition probabilities $\hat{p}_k((s' | (s,n),a))$ at the beginning of episode~$k$. 

At each time index, PUCRL2  checks the number of hits to (state, period index, action) tuples and state transitions. 
Like UCRL2, PUCRL2 proceeds in episodes. 
At the beginning of each episode, it  computes the estimates of the reward function and the transition probabilities from  past observations (Step 1). 
With high probability, the true AMDP lies within a confidence region computed around these estimates as shown in Lemma~\ref{lemma:probability-of-mdp-outside-confidence-region} (Step 2). Then PUCRL2 utilizes the confidence bounds as in \eqref{eqn:confidence_bound_p} and \eqref{eqn:confidence_bound_r}, to find an optimistic AMDP $\tilde{M}_k$ and policy $\tilde{\pi}_k$ using Modified-EVI Algorithm~\ref{alg:MEVI} adapted from the extended value iteration (EVI  \cite[Section 3.1.2]{auer2008near}) (Step 3). This policy $\tilde{\pi}_k$ is used to take action in the episode until the cumulative number of visits to any  (state, period index) pair, stored in $v_k((s,n),a)$, gets doubled; this is similar to the doubling criteria for episode termination of \cite{auer2008near} (Step 4).

\begin{algorithm}[t!]
\caption{PUCRL2}\label{alg:PUCRL2}
\begin{algorithmic}
\STATE \textbf{Input:} $S,A,N,$ confidence parameter $\delta  \in (0,1) $.\\
\STATE \textbf{Initialization:} $t=1, n = 1 $ \\
\FOR{episode \textit{k} = 1,2,...} 
\STATE{$t_k = t$ \COMMENT {starting time of episode k}}
\STATE{\textbf{1. Initialize episode} \textit{k}:} 
\footnotesize
$v_k((s,n),a)=0$, \\ $n_k((s,n),a) = max\{1,\sum_{\tau =1}^{t-1} {\mathds{1}_{\{((s_\tau,n_\tau),a_\tau) = ((s,n),a)\}}\}}$, \\ $n_k((s,n),a,s') = max\{1,\sum_{\tau = 1}^{t-1}{\mathds{1}_{\{((s_\tau,n_\tau),a_\tau,s_{\tau+1}) = ((s,n),a,s')\}}\}}$
\newline
\STATE $\hat{p}_k(s'|(s,n),a) = \frac{n_k((s,n),a,s')}{n_k((s,n),a)} \forall{(s,n),a} $ 
\newline
\STATE $ \hat{r}_k((s,n),a) = \frac{\sum_{\tau = 1}^{t-1}(r_\tau\mathds{1}_{((s_\tau,n_\tau),a_\tau) = ((s,n),a)} )}{n_k((s,n),a)} \forall{(s,n),a} $
\normalsize
\vspace{+2mm}
\newline
\STATE{\textbf{2. Update the confidence set}: We define the confidence region for  transition probability   and reward functions as:
\footnotesize
\begin{eqnarray} \label{eqn:confidence_bound_p}
&\mathcal{P}((s,n),a) \coloneqq \{ \mathbf{\Tilde{p}}(\cdot | (s,n),a) :  \nonumber \\ 
& \lVert \mathbf{\Tilde{p}}(\cdot | (s,n),a) - \mathbf{\hat{p}_k}((\cdot | (s,n),a)) \rVert_1 \leq \sqrt{\frac{14SN \log(2At_k / \delta)}{n_k((s,n),a)}} \}
\end{eqnarray} 
\begin{eqnarray} \label{eqn:confidence_bound_r}
&\mathcal{R}((s,n),a) \coloneqq \{ \Tilde{r}((s,n),a) : \nonumber \\  & \mid \Tilde{r}((s,n),a) - \hat{r}_k((s,n),a) \mid \leq \sqrt{\frac{7 \log(2SAt_k / \delta)}{2 n_k((s,n),a)}} \}
\end{eqnarray} 
\normalsize
Then, $\mathcal{M}_k$ is the set of all AMDP models, such that  \eqref{eqn:confidence_bound_p} and \eqref{eqn:confidence_bound_r} is satisfied for all $((s,n),a)$ pair.}
\STATE{\textbf{3. Optimistic Planning: Compute $(\tilde{M}_k,\tilde{\pi}_k) =$ Modified-Extended Value Iteration \ref{alg:MEVI}$(\mathcal{M}_k,\epsilon_k = 1/\sqrt{t_k})$}}
\newline
\STATE{\textbf{4. Execute Policies:}}
\WHILE{$v_k(n(s,n),a) < n_k((s,n),a)$}
\STATE Draw $a_t \sim \tilde{\pi}_k$; observe reward $r_t$ and   next state $s_{t+1}$.
 \STATE Set $v_k((s_t,n_t),a_t) = v_k((s_t,n_t),a_t) +1$ and $t =t+1,  n =((t-1)  \mod{N})+1)$ 
\ENDWHILE
\ENDFOR
\end{algorithmic}
\end{algorithm}

\subsection{Modified-EVI}

Extended value iteration is used in the class of UCRL algorithms to obtain an optimistic AMDP model and policy from a high probability confidence region.
According to the convergence criteria of Extended Value Iteration as in \cite[Section 3.1.3]{auer2008near},  aperiodicity is essential, i.e., the algorithm should not choose a policy with periodic transition matrix. 
However, the AMDP has a specific structure due to the periodicity of the original PMDP. Hence, in order to guarantee convergence, we modify the EVI algorithm by applying an aperiodicity transformation (as in \cite[Section 8.5.4]{puterman2014markov} ) \eqref{eqn:aperiodicity_transform}. At each iteration, Modified-EVI (Algorithm~\ref{alg:MEVI}) applies a self transition probability of $(1-\tau)$, where $0<\tau<1$, to the same (state, period index) pair. As shown in \cite[Proposition 8.5.8]{puterman2014markov}, this transformation does not affect the average reward of any stationary policy.

\begin{algorithm}[t!]
\caption{Modified - EVI}\label{alg:MEVI}
\begin{algorithmic}
\STATE \textbf{Input:} $ \mathcal{M}_k, \epsilon =  1/\sqrt{t_k}$\\
\STATE \textbf{Initialization:} $u_0(s,n) = 0 \forall s,n,s^{*} \in \mathcal{S}, n^{*} \in \{1,...N\}$ \\
\FOR{i = 0,1,2,...} 
\vspace{-5mm}
\STATE 
\footnotesize
\begin{equation} \label{eqn:aperiodicity_transform}
\begin{split}
u_{i+1}(s,n) & = \max_{a \in \mathcal{A}} \{ \max_{\dot{r} \in \mathcal{R}((s,n),a)} {\dot{r}((s,n),a)} \\
 & + \tau * \max_{\dot{p}  \in \mathcal{P}((s,n),a)} \{\sum_{s'} u_i(s',n+1)\dot{p}(s'|(s,n),a)\} \\
 & + (1-\tau) * u_i(s,n) 
\end{split}
\end{equation}

\STATE $u_{i+1}(s,n)  = u_{i+1}(s,n) - u_{i+1}(s^{*},n^{*})$
\vspace{+2mm}
\IF{ $\max_{(s,n)} \{ u_{i+1}(n,s) - u_{i}(n,s)\} - \min_{(s,n)} \{u_{i+1}(n,s) - u_{i}(n,s)\}  \leq \epsilon $}
\normalsize
\STATE Break the for loop.
\ENDIF
\ENDFOR
\end{algorithmic}
\end{algorithm}

\subsection{Analysis}
Let $M$ be a generic AMDP designated by the transition probabilities and reward functions. 
Let $T((s',n')| M, \pi, (s,1))$ denote the expected first hitting time of  $(s',n')$ for $M$, starting from $(s,1)$ under a stationary policy $\pi : \mathcal{S} \times \{1,2,....,N\} \xrightarrow{} \mathcal{A}$ . As in \cite[Definition 1]{auer2008near} the diameter of an AMDP $M$ is defined as:

\footnotesize
\begin{equation} \label{eqn:diameter}
    D_{aug} = \max _{(s',n')\neq (s,1),(s',s) \in \mathcal{S}^2 }\min_{\pi} \mathds{E}[T((s',n')| M, \pi,  (s,n))]
\end{equation}
\normalsize
\newtheorem{1}{}
\begin{theorem} \label{theorem:regret_bound}
With probability at least $1-\delta$,  the regret for PUCRL2 is:
\footnotesize
\begin{equation*}
\Delta(PUCRL2) \leq 34D_{aug} SN\sqrt{AT \log \frac{T}{\delta}}
\end{equation*}
\end{theorem}
\normalsize
\begin{proof}
See Appendix \ref{sec:proof_theorem_1}. 
\end{proof}

\subsection{PUCRLB algorithm}
In this section, we improve upon the previous algorithm by taking into account the special structure that arises out of augmentation of PMDP.
 Utilising the information about the periodicity of the transition matrix of the AMDP as discussed in Section~\ref{Problem Formulation}, we provide a modification of UCRLB algorithm, PUCRLB. Similar to \cite[Section 3.4]{fruit2019exploration}, we define:
 
 \footnotesize
 \begin{equation} \label{eqn:pucrlb_gamma}
     \Gamma^{\mathcal{S}}((s,n),a) = \lVert \mathbf{p}(\cdot | (s,n),a) \rVert_0 = \sum_{s'} \mathds{1}_{\{ p(s'|(s,n),a) >0 \}}
\end{equation}
\normalsize

Due to the periodic nature, the transition from any state-action pair $((s,n), a) \in  \mathcal{S} \times \mathcal{N} \times\mathcal{A}$ is limited to  $s' \in \mathcal{S}$, where the next period index is implicit by the previous one. This speciality is highlighted upon by  the superscript in \eqref{eqn:pucrlb_gamma}.

The PUCRLB algorithm is similar to PUCRL2. The main difference lies in the use of concentration inequalities which govern the construction of the set $\mathcal{M}_k$ of {\em candidate} AMDP's. While PUCRL2 uses Weisserman's \cite{weissman2003inequalities} and Hoeffding's inequalities to bound the $L_1$ norm of transition probability vector and reward function respectively, PUCRLB uses Empirical Bernstein Inequality \cite[Theorem 1]{audibert2009exploration} to bound the functions (Step 2). The transition function is bound individually for each $((s,n), a,s')$ pair, where $s'$ is an implicit representation of $(s',(n+1) \mod N)$. Thus, in the algorithm additionally we calculate the population variances of  reward and transition probabilities estimates, as:

\footnotesize
\begin{equation*}
\hat{\sigma}_{p,k}^2(s'|(s,n),a) = \hat{p}_k(s'|(s,n),a) (1-\hat{p}_k(s'|(s,n),a))
\end{equation*}
\begin{equation*}
\hat{\sigma}_{r,k}^2(s'|(s,n),a) = \frac{\sum\limits_{t=1}^{t_k-1} \mathds{1}_{\{((s_\tau,n_\tau),a_\tau) = ((s,n),a)\}} r_\tau^2}{n_k((s,n),a)} - (\hat{r}_k((s,n),a))^2 
\end{equation*}
\normalsize

Algorithm~\ref{alg:PUCRLB} details of all the changes necessary in Step 2 and Step 3 of  PUCRL2, that yield PUCRLB.

\newtheorem{2}{}
\begin{theorem} \label{theorem:pucrlb_regret_bound}
With probability at least $1-\delta$,  the regret for PUCRLB is:
\footnotesize
\begin{equation*}
\Delta(PUCRLB) \leq \beta D_{aug} \underbrace{S\sqrt{NAT\log(\frac{T}{\delta})}}_{\doteq \Delta_1} + \underbrace{D_{aug} S^2NA\log(\frac{T}{\delta}) \log(T)}_{\doteq \Delta_2}
\end{equation*}
\end{theorem}
\normalsize
\begin{proof}
See Appendix \ref{sec:proof_theorem_2}. 
\end{proof}

\begin{algorithm}[H]
\caption{PUCRLB (Modified Step-2,3 from PUCRL2)}\label{alg:PUCRLB}
\begin{algorithmic}
\STATE{\textbf{2. Update the confidence set}}: We define the confidence region for the transition probability function and reward functions as:
\vspace{-5mm}
\footnotesize
\begin{eqnarray} \label{eqn:pucrlb_confidence_bound_p}
& \mathcal{B}_p^k((s,n)a,s') \coloneqq \nonumber \\
& [ \hat{p}_k(s' |(s,n),a) - \beta_{p,k}^{(s,n),a,s'} , \hat{p}_k(s' |(s,n),a) + \beta_{p,k}^{(s,n),a,s'} ] \\ & \cap [0,1] 
\end{eqnarray} 
\begin{eqnarray} \label{eqn:pucrlb_confidence_bound_r}
& \mathcal{B}_r^k((s,n),a) \coloneqq \nonumber \\
& [ \hat{r}_k((s,n),a) - \beta_{r,k}^{(s,n),a} , \hat{r}_k((s,n),a) + \beta_{r,k}^{(s,n),a} ] \cap [0,1] 
\end{eqnarray} 
where
\footnotesize
\begin{eqnarray} \label{eqn:pucrlb_beta_p}
\begin{split}
  \beta_{p,k}^{(s,n),a,s'} & \coloneqq 2 \hat{\sigma}_{p,k}(s'|(s,n),a)\sqrt{\frac{\log(6SNA n_k/\delta)}{n_k((s,n),a)}} \\
  & + \frac{6\log(6SNA n_k/\delta)}{n_k((s,n),a)}   
\end{split}
\end{eqnarray}

\begin{eqnarray} \label{eqn:pucrlb_beta_r}
\begin{split}
\beta_{r,k}^{(s,n),a}  &\coloneqq 2 \hat{\sigma}_{r,k}((s,n),a)\sqrt{\frac{\log(6SNA n_k/\delta)}{n_k((s,n),a)}} \\
& + \frac{6\log(6SNA n_k/\delta)}{n_k((s,n),a)} 
\end{split}
\end{eqnarray}

\normalsize
Let $\mathcal{M}_k$ be the the set of all AMDP models coming from the confidence sets defined in \eqref{eqn:pucrlb_confidence_bound_p}   and \eqref{eqn:pucrlb_confidence_bound_r}.

\STATE{\textbf{3. Optimistic Planning:}} Compute $(\tilde{M}_k,\tilde{\pi}_k) =$ Modified-Extended Value Iteration \textbf{\ref{alg:MEVI}}$(\mathcal{M}_k,\epsilon_k = 1/t_k)$

\end{algorithmic}
\end{algorithm}

\subsection{Comparison between PUCRL2 and PUCRLB} \label{subsection:comparison}
We compare the regret bound obtained in Theorem \ref{theorem:regret_bound} and \ref{theorem:pucrlb_regret_bound} in terms of $\Tilde{O}$ (i.e. ignoring logarithmic terms). For $T\geq D_{aug}S^2NA$,
\footnotesize
\vspace{-1mm}
\begin{equation*}
\begin{split}
        \Delta(PUCRL2) & = \tilde{O}(D_{aug}SN\sqrt{AT})\geq \tilde{O}(S\sqrt{D_{aug}NAT})\\& \geq \tilde{O}(D_{aug}S^2NA) = \Delta_2.
\end{split}
\end{equation*}
\normalsize
Now, trivially 
\vspace{-1mm}
\footnotesize
\begin{equation*}
\begin{split}
        \Delta(PUCRL2) = \tilde{O}(D_{aug}SN\sqrt{AT}) \geq \tilde{O}(D_{aug}S\sqrt{NAT}) = \Delta_1.
\end{split}
\end{equation*}
\normalsize
\hspace{-1mm}
Thus PUCRLB yields a better regret bound than PUCRL2.

\section{Extended Uncertainty:   unknown period}
\label{section:unknown-N}

\begin{algorithm}[t!]
\caption{U-PUCRL2}\label{alg:Unknown N}
\begin{algorithmic}
\STATE \textbf{Input:} $S,A,$ confidence parameter $\delta  \in (0,1), $ set of candidate periods $\mathcal{N} = \{N_1, N_2, N_3,...N_l\}$\\
\STATE \textbf{Initialization:} $t=1,\hat{\rho}_{k,1} =0, n_i = 1 $ where, $ n_i \in \{1,2,...,N_i\} \forall i \in [l] \coloneqq  \{1,2, . . ,l\}$ \\
\STATE $\hat{p}_{1,i}(s'| (s,n_i),a) = 0 , \hat{r}_{1,i}((s,n_i),a) = 0$
\STATE $n_{1,i}((s,n_i),a) = 0, n_{1,i}((s,n_i),a,s') = 0$, for all ${(s,n_i),a,s'}, {n_i \in \{1,2,...,N_{i}\}},i \in [l] $
\FOR{episode \textit{k} = 1,2,...} 
\STATE{$t_k = t$} \text(starting time of episode k)
\STATE{\textbf{1. Initialize episode} \textit{k}:} 
\STATE For all ${(s,n_i),a}, {n_i \in \{1,2,...,N_{i}\}},i \in [l] $\\
\footnotesize
\STATE $v_{k,i}((s,n_i),a) =0,$ \\ 
\STATE $n_{k,i}((s,n_i),a) = max\{1,\sum_{\tau =1}^{t-1} {\mathds{1}_{\{((s_\tau,n_{i_\tau)},a_\tau) = ((s,n_i),a)\}}\}},$ \\ 
\vspace{-2mm}
\STATE
\begin{eqnarray*} 
&&n_{k,i}((s,n_i),a,s') \\
& =& max\{1,\sum_{\tau = 1}^{t-1}{\mathds{1}_{\{((s_\tau,n_{i_\tau)}),a_\tau,s_{\tau+1}) = ((s,n_i),a,s')\}}\}}
\end{eqnarray*} 
\STATE $\hat{p}_{k,i}(s'|(s,n_i),a) = \frac{n_k((s,n_i),a,s')}{n_k((s,n_i),a)}$ \\
\STATE $\hat{r}_{k,i}((s,n_i),a)  = \frac{\sum_{\tau = 1}^{t-1}(r_\tau\mathds{1}_{\{((s_\tau,n_{i_\tau)}),a_\tau) = ((s,n_i),a)} )\}}{n_{k,i}((s,n_i),a)}$
\normalsize
\newline
\STATE{\textbf{2. Calculate estimated average reward}}:
\STATE $\hat{\rho}_{k,i} = \hat{\rho}_{k-1,i} + Value  Iteration(\mathbf{\hat{p}_{k,i}},\mathbf{\hat{r}_{k,i})}$
\STATE {\textbf{3. Choose the period with highest value}} :
\STATE $I_k = \argmax_i \hat{\rho}_{k,i}$

\STATE{\textbf{4. Update the confidence set}}: We define the confidence region for  transition probability function and reward functions as:
\vspace{-2mm}
\footnotesize
\begin{eqnarray*} 
&\mathcal{P}((s,n_{I_k}),a) \coloneqq \{ \mathbf{\Tilde{p}}(\cdot | (s,n_{I_k}),a) :  \nonumber \\ 
& \lVert \mathbf{\Tilde{p}}(\cdot | (s,n_{I_k}),a) - \mathbf{\hat{p}_{k,I_k}}((\cdot | (s,n_{I_k}),a)) \rVert_1 \\
&\leq \sqrt{\frac{14SN_{I_k} \log(2At_k / \delta)}{n_{k,I_k}((s,n_{I_k}),a)}} \}
\end{eqnarray*} 
\begin{eqnarray*}
&\mathcal{R}((s,n_{I_k}),a) \coloneqq \{ \Tilde{r}((s,n_{I_k}),a) : \nonumber \\  & \mid \Tilde{r}((s,n_{I_k}),a) - \hat{r}_{k,I_k}((s,n_{I_k}),a) \mid \leq \sqrt{\frac{7 \log(2SAt_k / \delta)}{2 n_{k,I_k}((s,n_{I_k}),a)}} \}
\end{eqnarray*} 
\normalsize
Then, $\mathcal{M}_{k,{I_k}}$ is the set of all MDP models, such that above equations are satisfied for all $((s,n_{I_k}),a)$ tuples for all $n_{I_k} \in \{1,2,...,N_{I_k}\}$.
\STATE{\textbf{5. Optimistic Planning: Compute $(\tilde{M}_{k,{I_k}},\tilde{\pi}_{k,{I_k}}) =$ Modified-EVI $(\mathcal{M}_{k,{I_k}},1/\sqrt{t_k})$}}
\newline
\vspace{-2mm}
\STATE{\textbf{6. Execute Policies:}}
\WHILE{$v_k((s_t,n_{I_k}),a_t) < n_k((s_t,n_{I_k}),a_t)$}

\STATE Draw $a_t$ according to $\tilde{\pi}_{k,I_k}$, observe reward $r_t$ and the next state $s_{t+1}$.
 \STATE Set $v_{k,i}((s_t,n_{i}),a_t) = v_{k,i}((s_t,n_{i}),a_t) +1 $ and $t =t+1,n_i =((t-1)\mod{N_i})+1) \hspace{+2mm} \forall i \in [l]$ 
\ENDWHILE
\ENDFOR
\end{algorithmic}
\end{algorithm}
In this section, we consider the scenario where $N$ is unknown. However, we assume a set of candidate periods $\mathcal{N} = \{N_1, N_2, N_3,...N_l\}$   which contains the true period $N$. This setup demands extra exploration from the agent to identify the true period with high accuracy which can be used to then model the environment and perform exploitation. 
We provide an alternative algorithm  Unknown-PUCRL2 or U-PUCRL2  (Algorithm~\ref{alg:Unknown N}) for learning, which is an extension of PUCRL2. The reward function $\hat{r}_{k,i}((s,n_i),a)$ and transition function $\hat{p}_{k,i}(s'|(s,n_i),a)$ estimates are maintained for each candidate period (denoted by subscript $i$) separately considering their period information is true and using it to calculate respective period indices at each time step (Step 1).

 At the beginning of each episode $k$, these estimates are used to calculate an estimate of average reward through Value Iteration algorithm \cite[Algorithm 8.5.1]{puterman2014markov}, for each candidate period $N_i, i \in [l]$. Based on the hypothesis that the true candidate period will have the true representation of the underlying AMDP and hence will have the highest average reward, the candidate period with the highest cumulative average reward is selected as the true period for that episode (Step 3).

  Based on the selected period information, policy for that episode is calculated through Algorithm~\ref{alg:MEVI}. The observation tuple ($s_t,a_t,r_t,s_{t+1}$) is used to update the estimate for every candidate period $N_i, i \in [l]$ (Step 6). This is valid since the underlying AMDP would produce the same tuple even if some other candidate's policy would have selected the same action in that state. 
  
  {\em U-PUCRLB:} In a similar way, we can also design U-PUCRLB for unknown $N$. However, its details are omitted in this paper for brevity.

\section{Numerical results}
\label{section:numerical}
We compare the performance of all the aforementioned algorithms with other state of the art algorithms: (i) UCRL2 \cite{auer2008near} which provides optimal static regret in stationary MDP setting, (ii) UCRL3 \cite{bourel2020tightening} which is a recent improvement over UCRL2, (iii) BORL \cite{cheung2020reinforcement} which is a   parameter free algorithm for the non-stationary setting, (iv) PSRL \cite{osband2013more}, an adaption of Thomson Sampling to RL.

\subsection{Regret of BORL for PMDP}
The variation budget \cite{cheung2020reinforcement} for the rewards is defined as $
    B_r = \sum_{t=1}^{T-1} \max_{s \in \mathcal{S}, a \in \mathcal{A}}| r_{t+1}(s,a) - r_t(s,a) |$. 
For a PMDP:
\vspace{-1mm}
\footnotesize
\begin{equation*} 
\begin{split}
B_r & = \sum_{t=1}^{T-1} \max_{s \in \mathcal{S}, a \in \mathcal{A}}| r_{t+1}(s,a) - r_t(s,a) |  \\
 &  \approx (T/N) \sum_{t=1}^N \max_{s \in \mathcal{S}, a \in \mathcal{A}}| r_{t+1}(s,a) - r_t(s,a) | \approx \mathcal{\Tilde{O}}(T)
\end{split}
\end{equation*}

\normalsize

Regret bounds of BORL and SW-UCRL \cite{cheung2020reinforcement} for non-stationary MDP are derived in terms of the reward variation budget $B_r$ and a very similar variation budget $B_p$ on the transition kernels. However, for a PMDP, these two algorithms do not exploit the additional structure arising out of periodicity. Since $B_r$ or $B_p$ turn out to be of the order $\mathcal{\Tilde{O}}(T)$ , the $\mathcal{\Tilde{O}}((B_r+B_p)^{1/4}T^{3/4})$ regret bound of  BORL or SW-UCRL becomes $\mathcal{\Tilde{O}}(T)$ for PMDP.

\begin{figure} 
  \centering
  \includegraphics[scale=0.5]{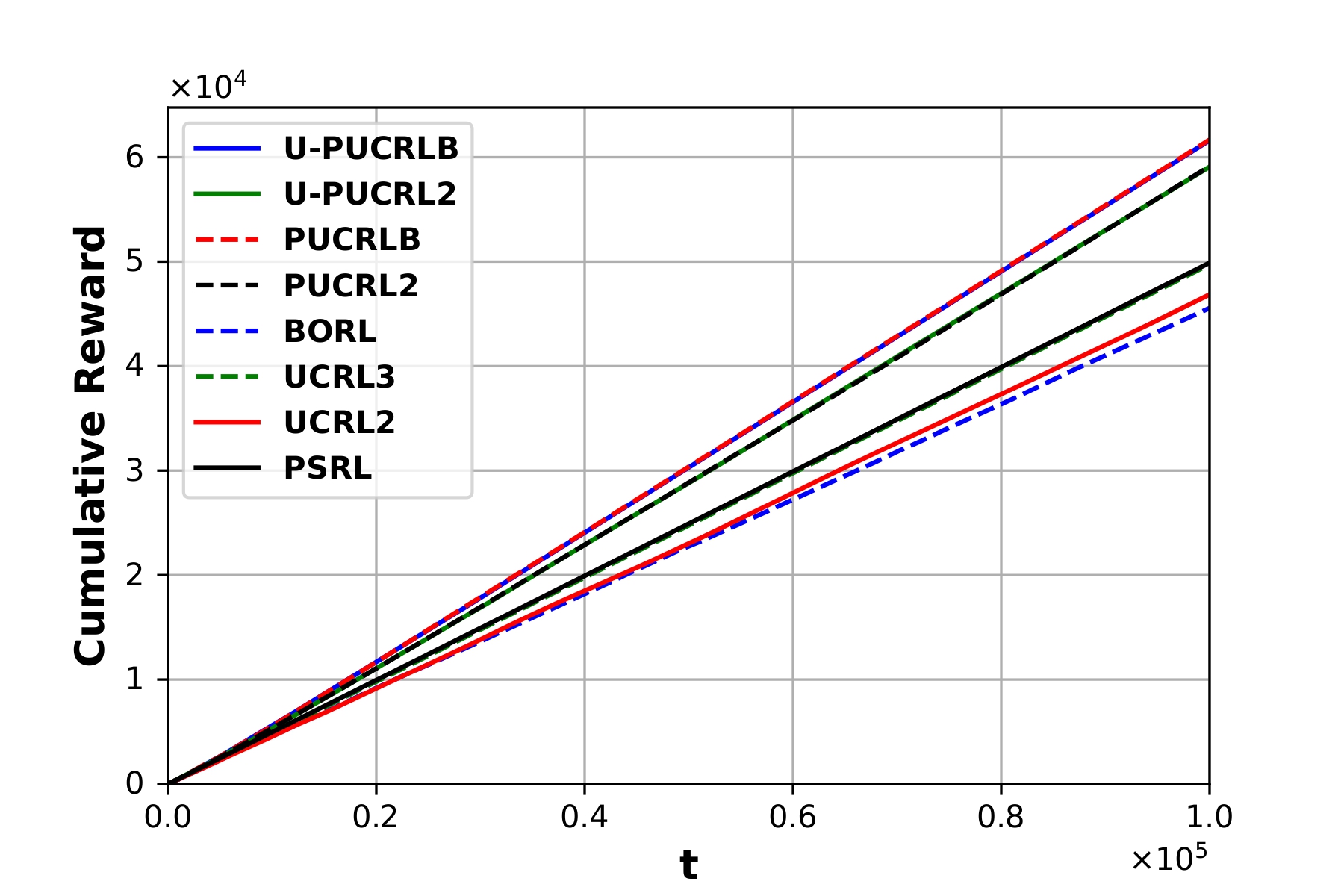}
  \includegraphics[scale=0.5]{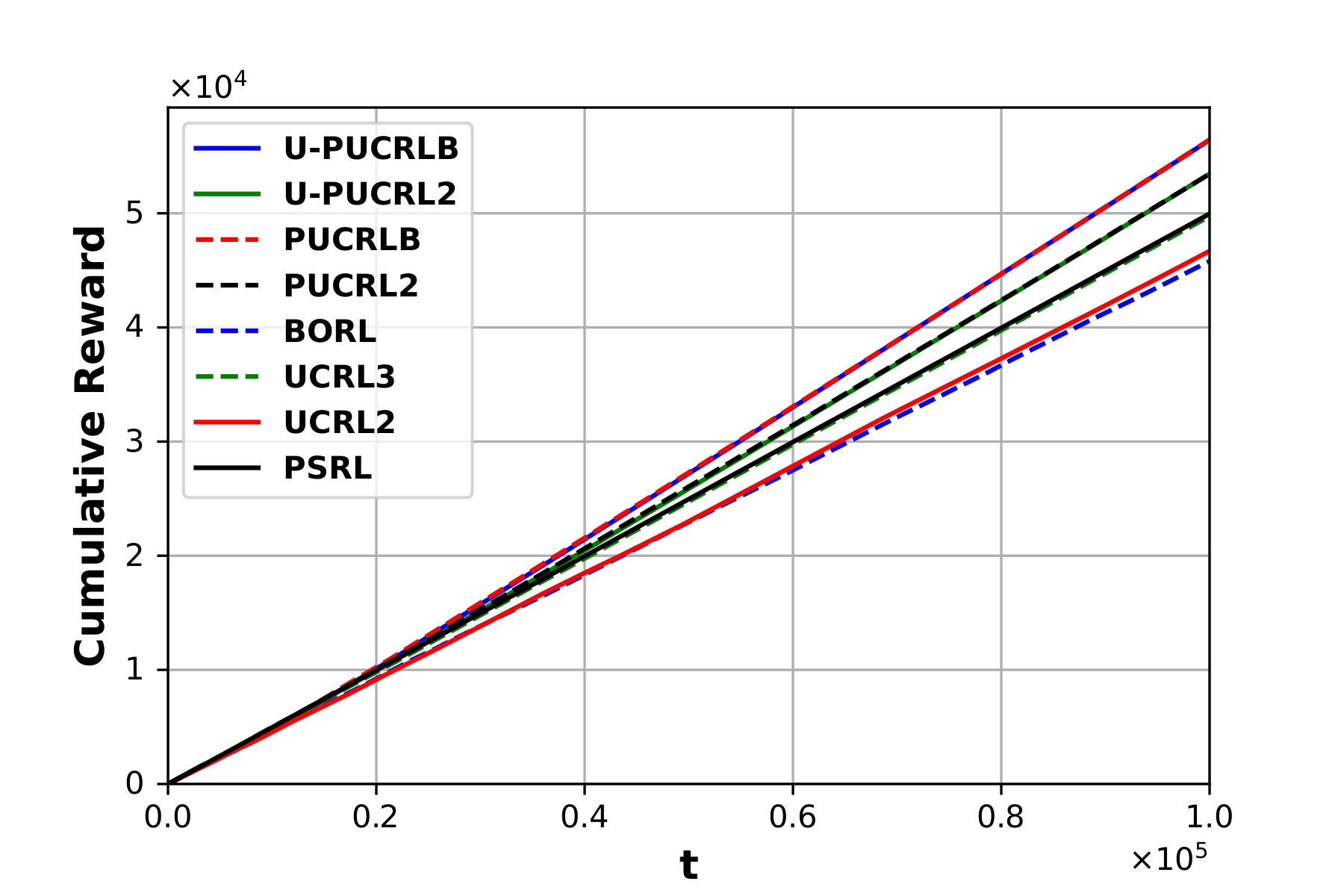}
  \vspace{-5mm}
  \caption{Cumulative reward for a  2-state, 2-action PMDP with N = 5 (Above) and  N = 15 (Below).}
  \label{fig:Reward_plot}
  \vspace{-5mm}
\end{figure}

\begin{figure} 
  \centering
  \includegraphics[scale=0.5]{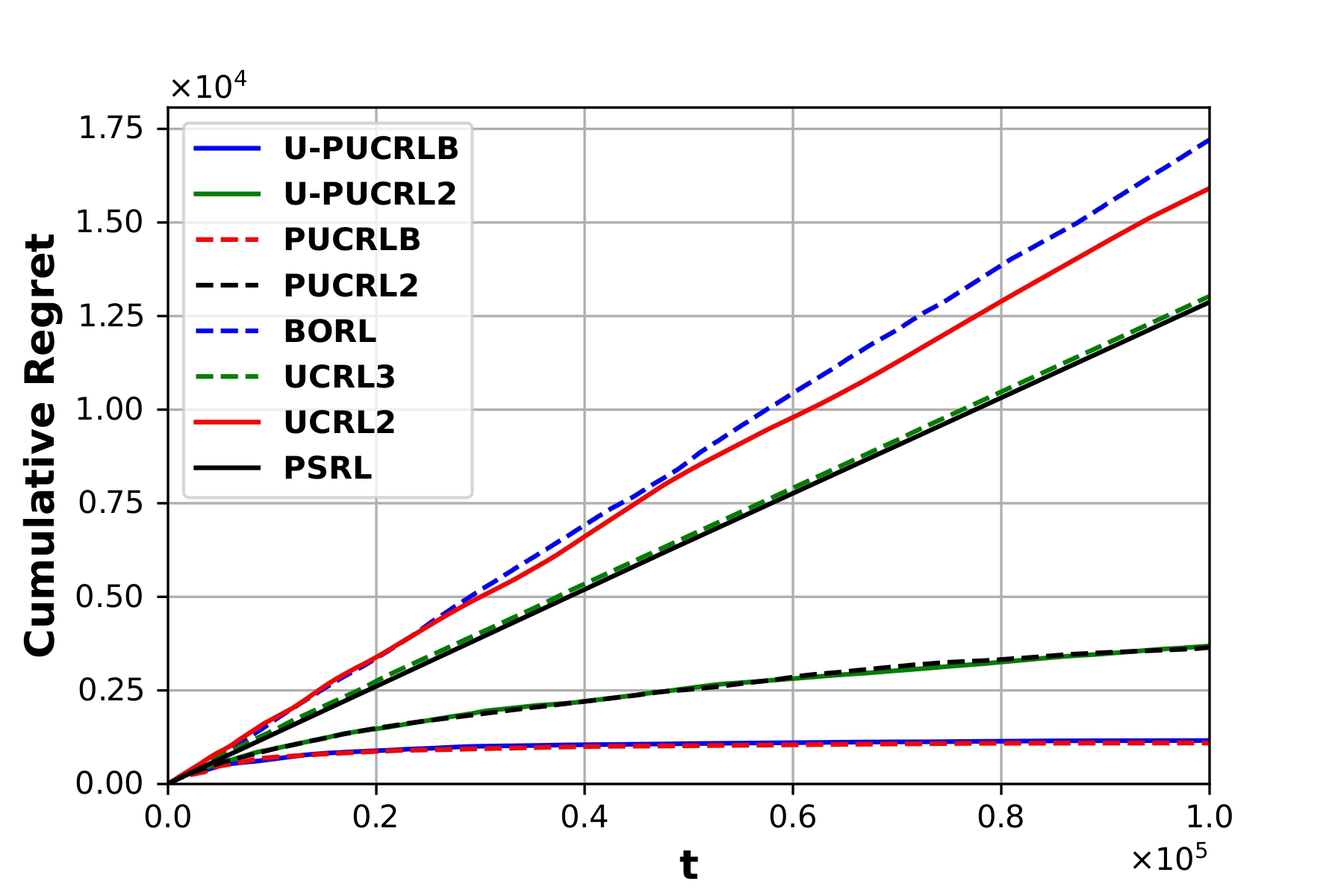}
  \includegraphics[scale=0.5]{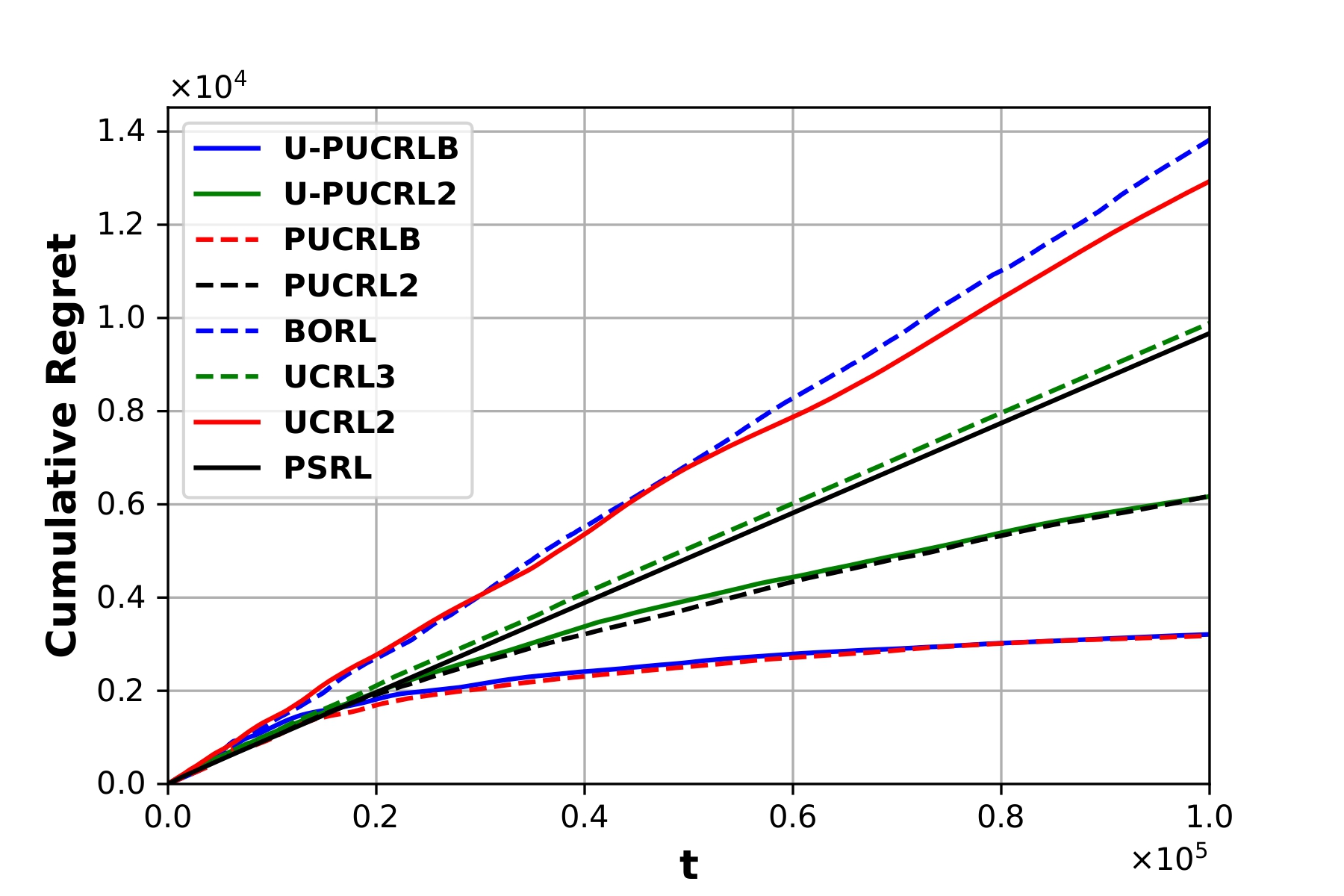}
  \caption{Cumulative Regret with N = 5 (Above) and  N = 15 (Below).}
  \label{fig:Regret_plot}
\end{figure}

\subsection{Our experiment}
We perform empirical analysis on synthetic data-set. We consider a MDP with two states $\{s_1, s_2\}$, two actions  $\{a_1, a_2\}$ and $T=  100000$. The variation in the rewards and transition function are modeled using saw-tooth functions as follows:

\vspace{-2mm}
\footnotesize
\begin{equation*} 
\begin{split}
r_t(s_1,a_1) & = 0.5 + \arctan(1 / \tan(\pi*(t+0.5)/N)) /N \\
r_t(s_1,a_2) & = 0.5 - \arctan(1 / \tan(\pi*(t+0.5)/N)) /N \\
r_t(s_2,a_1) & = 0.4 + 0.8 *(t/N - floor(0.5+t/N)) \\r_t(s_2,a_2) & = 0.4 - 0.8 *(t/N - floor(0.5+t/N))
\end{split}
\end{equation*}
\normalsize
\vspace{-5mm}
\footnotesize
\begin{eqnarray*}
p_t(s_1|s_1,a_1) &=& 1,  p_t(s_2|s_1,a_1) = 0,\\
p_t(s_1|s_1,a_2) &=& 1-\beta_t,  p_t(s_2|s_1, a_2) = \beta_t, \\
p_t(s_1|s_2,a_1) &=& 0,  p_t(s_2|s_2,a_1) = 1,  \\
p_t(s_1|s_2,a_2) &=& \beta_t,  p_t(s_2|s_2, a_2) = 1-\beta_t
\end{eqnarray*}

\normalsize
where,
$\beta_t = 0.5 - \arctan(1 / \tan(\pi*(t+0.5)/N)) /N$. We set the period $N = 5 $ and $15$, the candidate period sets $\mathcal{N} = \{2,3,4,5,6,7\}$ and $ \{12,13,14,15,16,17,18\}$, $\delta=0.05$,  and  compare the cumulative reward of the algorithms after averaging   over
$30$ independent runs. Figure~ \ref{fig:Regret_plot} depicts the cumulative regret  and Figure~\ref{fig:Reward_plot} shows cumulative reward accrued by different algorithms over the time horizon respectfully. 
We clearly observe that our algorithms outperform other algorithms. Specifically, PUCRLB performs the best as discussed in Section~\ref{subsection:comparison}. We also notice that PUCRL2 and U-PUCRL2 have similar performance because, U-PUCRL2 learns the true $N$ and then behaves like PUCRL2.  The same can be observed in PUCRLB and   U-PUCRLB.

\section{Conclusion}
In this paper, we have studied periodic non-stationarity in Markov
Decision Processes, where the state transition and reward
functions vary periodically. Existing RL algorithms for non-stationary and stationary MDPs fail to perform optimally in this setting. We have proposed two algorithms called PUCRL2 and PUCRLB, which outperform competing algorithms. We have also extended the uncertainty in the already varying environment by considering   unknown period, and have shown  numerically that  lack of knowledge of period does not matter to the long-term reward and regret performance. However,  
the static regret term depends linearly on the diameter of the AMDP, the characterization of which with $N$  is still open.

{\small
\bibliographystyle{unsrt}
\bibliography{reference.bib}
}

\newpage
 \appendices

 \section{PROOF OF THEOREM 1}
 \label{sec:proof_theorem_1}

The proof borrows some ideas from \cite{auer2008near} and is divided into sections. 
In Appendix \ref{subsec:splitting_into_episodes}, we upper bound the total regret by removing the randomness in the rewards accumulated. The regret in the episodes where the true AMDP does not lie in the set of plausible AMDPs is bounded above in Appendix \ref{subsec:failing_confidence_region}, and with the assumption that it does in Appendix \ref{subsec:M_in_M_k}. Finally, we complete the proof in Appendix \ref{subsec:completing_the _proof}.

\subsection{Splitting into episodes} \label{subsec:splitting_into_episodes}

As in \cite[Section 4.1]{auer2008near} using Hoeffding’s inequality , we can decompose the regret as:
\footnotesize
\begin{eqnarray*} \label{eqn2}
\Delta &=& \sum_{t =1}^{T} (\rho^* - r_t((s_t,n_t),a_t)) \\
  &\leq& T\rho^* - \sum_{(s,n),a} N((s,n),a) r((s,n),a) + \sqrt{\frac{5}{8}T \log \frac{8T}{\delta}}  
\end{eqnarray*}
\normalsize
with probability at least $1-\frac{\delta}{12 T^{5/4}}$
, where $N((s,n),a)$ is the count of (state, period)-action pair after $T$ steps.

Let there be m episodes in total , thus
$\sum_{k=1}^m v_k((s,n),a) = N((s,n),a)$.

The regret in each episode can be defined as : $\Delta_k = \sum_{(s,n),a} v_k((s,n),a) (\rho^* - r((s,n),a))$. Hence,
\footnotesize
\begin{equation} \label{eqn:regret_decomposition}
\Delta\leq \sum_{k=1}^m \Delta_k + \sqrt{\frac{5}{8}T \log \frac{8T}{\delta}}
\end{equation}
\normalsize

\subsection{ Dealing with failing confidence regions} \label{subsec:failing_confidence_region}

\begin{lemma} \label{lemma:probability-of-mdp-outside-confidence-region}
For any $t \geq 1$, the probability that the true AMDP M is not contained in the set of plausible
AMDPs $\mathcal{M}(t)$ at time t is at most $\delta / 15 t^6$, that
is 
\begin{equation*}
\mathbb{P}\{M \notin \mathcal{M}(t)\} < \delta / 15t^6
\end{equation*}

\end{lemma}
\begin{proof}
As in \cite[Section C.1]{auer2008near} we bound the transition functions using $L^1$-deviation concentration inequality over $m$ distinct events from $l$ samples \cite{weissman2003inequalities}:
\begin{eqnarray*}
\mathbb{P} \{ \lVert  \hat{\mathbf{p}}(\cdot) - \mathbf{p(\cdot)} \rVert_1 \geq \epsilon_p \} \leq (2^m-2) \exp(-l\epsilon_p^2 / 2)
\end{eqnarray*}

As the state space has been augmented, we have $SN$ states and hence $m = SN$ events. 
\newline
Thus, setting 
\footnotesize
\begin{eqnarray*}
\epsilon_p = \sqrt{\frac{2}{l} \log(\frac{2^{SN}20SAt^7}{\delta})} \leq \sqrt{\frac{14SN}{l} \log(\frac{2At}{\delta})}
\end{eqnarray*}
\normalsize
we get,
\footnotesize
\begin{eqnarray*}
\mathbb{P} \{ \lVert  \hat{\mathbf{p}}(\cdot|(s,n),a) - \mathbf{p}(\cdot|(s,n),a) \rVert_1 \geq \sqrt{\frac{14SN \log(2At / \delta)}{l}} \} \leq \frac{\delta}{20t^7SA}
\end{eqnarray*}
\normalsize
For rewards, we use Hoeffding's inequality to bound the deviation of empirical mean from true mean given $l$ i.i.d samples
\footnotesize
\begin{eqnarray*}
\mathbb{P} \{ \lvert  \hat{r} - r \rvert \geq \epsilon_r\} \leq 2 \exp(-2l\epsilon_r^2)
\end{eqnarray*}
\normalsize
Setting 
\footnotesize
\begin{eqnarray*}
\epsilon_r = \sqrt{\frac{1}{2l} \log(\frac{120SAt^7}{\delta})} \leq \sqrt{\frac{7}{2l} \log(\frac{2SAt}{\delta})}
\end{eqnarray*}
\normalsize
we get for all $((s,n),a)$ pair
\footnotesize
\begin{eqnarray*}
\mathbb{P} \{ |  \hat{r}((s,n),a) - r((s,n),a) | \geq \sqrt{\frac{7 \log(2SAt / \delta)}{2l}} \} \leq \frac{\delta}{60t^7SA}
\end{eqnarray*}

\normalsize
 A union bound over all possible values of $l$ i.e. $l$ = 1,2,..... $\lfloor t/N \rfloor$, gives ($n_k((s,n),a)$ denotes the number of visits in $((s,n),a)$)
\footnotesize

\begin{align}
\mathbb{P} \{ \lVert \hat{\mathbf{p}}(\cdot|(s,n),a) & - \mathbf{p}(\cdot|(s,n),a) \rVert_1 \geq \sqrt{\frac{14SN \log(2At / \delta)}{n_k((s,n),a)}} \} \nonumber \\
&\leq \sum_{t=1} ^{\lfloor t/N \rfloor} \frac{\delta}{20t^7SA} \nonumber \leq  \sum_{t=1} ^{t/N}  \frac{\delta}{20t^7SA}= \frac{\delta}{20t^6SAN} \nonumber
\end{align}
\begin{eqnarray*}
\mathbb{P} \{ |  \hat{r}((s,n),a) - r((s,n),a) | \geq \sqrt{\frac{7 \log(2SAt / \delta)}{2n_k((s,n),a)}} \} \leq \sum_{t=1} ^{\lfloor t/N \rfloor} \frac{\delta}{60t^7SA} \\
\leq \sum_{t=1} ^{t/N}  \frac{\delta}{60t^7SA}= \frac{\delta}{60t^6SAN}
\end{eqnarray*}

\normalsize

Summing these probabilities over all (state, period)-action pairs we obtain the claimed bound 
$\mathbb{P}\{M \notin \mathcal{M}(t)\} < \delta / 15t^6$.

\end{proof}

\begin{lemma} \label{lemma:regret_due_to_failing_confidence_region}
With probability at least $1-\frac{\delta}{12T^{5/4} }$, the regret occurred due to failing confidence region i.e. 
\begin{equation} \label{eqn:M_not_in_confidence_bound}
\sum_{k=1}^m \Delta_k \mathds{1}_{\{M \notin \mathcal{M}_k\}} \leq \sqrt{T}
\end{equation}
\end{lemma}
\begin{proof}
Refer \cite[Section 4.2]{auer2008near} with Lemma \ref{lemma:probability-of-mdp-outside-confidence-region} instead of \cite[Appendix C.1]{auer2008near}
\end{proof}

\subsection{Episodes with $M \in \mathcal{M}_k$} \label{subsec:M_in_M_k}

By the assumption $M \in \mathcal{M}_k$ and \cite[Theorem 7]{auer2008near}, the optimistic optimal average reward of the near optimal policy $\tilde{\pi}_k$ chosen in Modified-EVI \ref{alg:MEVI} is such that $\Tilde{\rho_k} \geq \rho^* - \epsilon_k$.

Thus, substituting $\epsilon_k = 1/\sqrt{t_k}$, we can write the regret of an episode as :

\footnotesize
\begin{equation} \label{eqn:epsiodic_regret_optimism}
\begin{split}
\Delta_k & = \sum_{(s,n),a} v_k((s,n),a) (\rho^* - r((s,n),a))  \\
 & \leq \sum_{(s,n),a} v_k((s,n),a) (\tilde{\rho}_k - r((s,n),a)) + \sum_{(s,n),a} \frac{v_k((s,n),a)}{\sqrt{t_k}}. 
\end{split}
\end{equation}

\normalsize
Let us define $i_k$ to be the last iteration when convergence criteria holds and Modified-EVI terminates, thus as in \cite[Section 4.3.1]{auer2008near}

\footnotesize
\begin{equation} \label{eqn:EVI_condition}
|u_{i_k+1}(s,n) - u_{i_k}(s,n) - \tilde{\rho_k}| \leq 1/\sqrt{t_k}  
\end{equation}
\normalsize
for all $(s,n)$. Expanding as in \eqref{eqn:aperiodicity_transform}
\footnotesize
\begin{eqnarray*} 
u_{i_k+1}(s,n) &=& \tilde{r}_k((s,n),\tilde{\pi}_k(s,n)) \\
 &+& \tau * \{\sum_{s'} u_{i_k}(s',n+1)\tilde{p}_k(s'|(s,n),\tilde{\pi}_k(s,n))\} \\
 &+& (1-\tau) * u_{i_k}(s,n)\}
\end{eqnarray*}

\normalsize

Putting it in \eqref{eqn:EVI_condition}, we get

\footnotesize
\begin{eqnarray*}  
|\tilde{\rho_k} - \tilde{r}_k((s,n),\tilde{\pi}_k(s,n)) 
- \tau * \{\sum_{s'} u_{i_k}(s',n+1)\tilde{p}_k(s'|(s,n),\tilde{\pi}_k(s,n))\} \\
- (\cancel{1}-\tau) * u_{i_k}(s,n)\ + \cancel{u_{i_k}(s,n)} | \leq 1/\sqrt{t_k} 
\end{eqnarray*}
\vspace{-5mm}
\begin{equation*} 
\begin{split}
\tilde{\rho_k}  - \tilde{r}_k((s,n),\tilde{\pi}_k(s,n)) & \leq  \tau * \{\sum_{s'} u_{i_k}(s',n+1) \\
 & \tilde{p}_k(s'|(s,n),\tilde{\pi}_k(s,n))\} - \tau * u_{i_k}(s,n) + 1/\sqrt{t_k}
\end{split}
\end{equation*}

\normalsize 
Thus, putting the above result in \eqref{eqn:epsiodic_regret_optimism}, and noting that $\sum_{(s,n),a} v_k((s,n),a) = 0$, for $a \neq \tilde{\pi}_k(s,n)$, we get

\footnotesize
\begin{equation} \label{eqn:del_p_del_r}
\begin{split}
\Delta_{k} & \leq \underbrace{\tau \sum_{(s,n),a} v_k((s,n),a) ( \sum_{s'} u_{i_k}(s',n+1)\tilde{p}_k(s'|(s,n),a)  -  u_{i_k}(s,n))}_{ \coloneqq \Delta_{k}^{p}} \\
& + \underbrace{\sum_{(s,n),a} v_k((s,n),a) (\tilde{r}_k((s,n),a)) - r((s,n),a))}_{\coloneqq \Delta_{k}^{r}} \\
& + 2 \sum_{(s,n),a} \frac{v_k((s,n),a)}{\sqrt{t_k}}  
\end{split}
\end{equation}

\normalsize

\subsubsection{Bounding $\Delta_k^p$} \label{sub_sub_Sec:del_p}

\footnotesize

\begin{equation} \label{eqn:del_p}
\begin{split}
\Delta_k^p & =  \tau\sum_{(s,n),a} v_k((s,n),a)( \{\sum_{s'} u_{i_k}(s',n+1)\tilde{p}_k(s'|(s,n),a)\} 
\\ & -  u_{i_k}(s,n))) \\
& = \tau\sum_{(s,n),a} v_k((s,n),a)( \sum_{s'} u_{i_k}(s',n+1) \\
& (\tilde{p}_k(s'|(s,n),a) - p_k(s'|(s,n),a)) + \tau\sum_{(s,n),a} v_k((s,n),a) 
\\  & ( \sum_{s'}u_{i_k}(s',n+1) p_k(s'|(s,n),a) - u_{i_k}(s,n)) 
\end{split}
\end{equation}

\normalsize
where $p_k(s'|(s,n),a)$ is the true transition probability (in M) of the policy applied in episode k for the tuple $((s,n),a,s')$ . By the property of extended value iteration\cite[Section 4.3.1]{auer2008near}, extended to Modified-EVI
\begin{equation} \label{eqn:span_bound}
     span(\mathbf{u}_{i_k}) = \max_{(s,n)} u_{i_k}(s,n) - \min_{(s,n)} u_{i_k}(s,n) \leq D_{aug}^\tau  \\
\end{equation} 
where $D_{aug}^\tau$ represents  the diameter of the augmented MDP with aperiodicity transformation.

Since, $ \sum_{s'} p_k(s'|(s,n),a) =1$ and $\sum_{s'}\tilde{p}_k(s'|(s,n),a)=1$, we can replace $u_{i_k}(s,n)$ by 

\footnotesize
\begin{equation} 
    w_k(s,n) = u_{i_k}(s,n) - \frac{\max_{(s,n)} u_{i_k}(s,n) + \min_{(s,n)} u_{i_k}(s,n)}{2}
\end{equation}
\normalsize
such that it follows from \eqref{eqn:span_bound} that $span(\mathbf{u}_{i_k}) = span (\mathbf{w}_k$).

Hence, $\lVert \mathbf{w}_k \rVert_\infty \leq D_{aug}^\tau/2$.

According to \cite[Section 3.3.1]{fruit2019exploration}, $D_{aug}^\tau  \leq D_{aug}/\tau$. Hence,
$\lVert \mathbf{w}_k \rVert_\infty \leq D_{aug}/2\tau$.

Thus, the first term in \eqref{eqn:del_p} can be bounded as :
 \footnotesize
\begin{equation*} 
{\tau}\sum_{(s,n),a} v_k((s,n),a)( \sum_{s'} w_k(s',n+1)(\tilde{p}_k(s'|(s,n),a) - p_k(s'|(s,n),a))
\end{equation*}
\begin{align*} 
\leq \tau\sum_{(s,n),a} v_k((s,n),a)(  \lVert \mathbf{w}_k \rVert_\infty  \lVert  \mathbf{\tilde{p}_k}(\cdot|(s,n),a) - \mathbf{p_k}(\cdot|(s,n),a) \rVert_1)
\end{align*}

\begin{align} \label{eqn:del_p_first_term}
\leq \sum_{(s,n),a} v_k((s,n),a) \cancel{2\tau}\sum_{(s,n),a} \sqrt{\frac{14SN \log(2At_k / \delta)}{n_k((s,n),a)}} D_{aug}/\cancel{2\tau}
\end{align}
\normalsize
where the last inequality uses the confidence bound \eqref{eqn:confidence_bound_p}. We note that the aperiodicity transformation coefficient gets canceled out and does not appear in the regret term.

Following the proof of \cite[Second term, Section 4.3.2]{auer2008near}, the second term in \eqref{eqn:del_p} can be bounded as:

\footnotesize
\begin{equation} \label{eqn:del_p_second_term}
\begin{split}
\tau \sum_{k=1}^m \sum_{(s,n),a} v_k((s,n),a) & ( \sum_{s'} u_{i_k}(s',n+1) p_k(s'|(s,n),a) - u_{i_k}(s,n))\\
& \leq \tau\ D_{aug}^\tau \sqrt{\frac{5}{2}T \log \frac{8T}{\delta}} + m \tau\ D_{aug}^\tau \\
& \leq \cancel{\tau}\ D_{aug}/\cancel{\tau} \sqrt{\frac{5}{2}T \log \frac{8T}{\delta}} + m \cancel{\tau} D_{aug}/\cancel{\tau} 
\end{split}
\end{equation}
\normalsize
with probability at least $1-\frac{\delta}{12T^{5/4}}$, where $m \leq SNA \log \frac{8T}{SNA}$ is the number of episodes as in \cite[Appendix C.2]{auer2008near}.
\vspace{+2mm}
\subsubsection{Bounding $\Delta_k^r$} \label{sub_sub_Sec:del_r}
\footnotesize
\begin{equation} \label{eqn:del_r}
\begin{split}
\Delta_k^r & = \sum_{(s,n),a} v_k((s,n),a) (\tilde{r}_k((s,n),a)) - r((s,n),a)) \\
& \leq \sum_{(s,n),a} v_k((s,n),a) (|\tilde{r}_k((s,n),a)) - \hat{r}_k((s,n),a))| \\
& + |\hat{r}_k((s,n),a)) - r((s,n),a))|) 
\\  & \leq 2\sum_{(s,n),a} v_k((s,n),a) \sqrt{\frac{7 \log(2SAt_k / \delta)}{2 n_k((s,n),a)}}   
\end{split}
\end{equation}

\normalsize
where the last inequality uses the confidence bound \eqref{eqn:confidence_bound_r}.

\subsection{Completing the Proof} \label{subsec:completing_the _proof}

Thus, we can write the total episodic regret using \eqref{eqn:del_p_del_r}, \eqref{eqn:del_p_first_term},\eqref{eqn:del_p_second_term}, and \eqref{eqn:del_r}, with probability at least $1-\frac{\delta}{12T^{5/4}}$:

\footnotesize
\begin{equation*} 
\begin{split}
\sum_{k=1}^m \Delta_k \mathds{1}_{\{M \in \mathcal{M}_k\}} & \leq \sum_{k=1}^m  \sum_{(s,n),a} v_k((s,n),a)D_{aug} \sqrt{\frac{14SN \log(2At_k / \delta)}{n_k((s,n),a)}} \\
& + D_{aug}\sqrt{\frac{5}{2}T \log \frac{8T}{\delta}} + D_{aug}  SNA \log \frac{8T}{SNA}
\\  & + (\sqrt{14 \log(2SAt_k / \delta)} + 2) \sum_{k=1}^m \sum_{(s,n),a} \frac{v_k((s,n),a)}{\sqrt{n_k((s,n),a)}} 
\end{split}
\end{equation*}

\normalsize
We can bound the term $ \sum_{k=1}^m \sum_{(s,n),a} \frac{v_k((s,n),a)}{\sqrt{n_k((s,n),a)}} \leq (\sqrt{2}+1)(\sqrt{SNAT})$ as in \cite[Section 4.3.3]{auer2008near}. Also, noting that $n_k((s,n),a)\leq t_k\leq T$.Thus,

\footnotesize
\begin{equation} \label{eqn:episodic_regret_bound}
\begin{split}
\sum_{k=1}^m \Delta_k \mathds{1}_{\{M \in \mathcal{M}_k\}} & \leq  D_{aug}\sqrt{\frac{5}{2}T \log \frac{8T}{\delta}} +   D_{aug}  SNA \log \frac{8T}{SNA} \\
& +(2D_{aug}\sqrt{14SN  \log(2AT / \delta)} + 2) \\& (\sqrt{2}+1)(\sqrt{SNAT})
\end{split}
\end{equation}
\normalsize

Using \eqref{eqn:regret_decomposition}, \eqref{eqn:M_not_in_confidence_bound},  \eqref{eqn:episodic_regret_bound},  with a probability of $1-\frac{\delta}{4T^{5/4}}$, we can bound the total regret as:

\footnotesize
\begin{equation*} 
\begin{split}
\Delta & \leq \sum_{k=1}^m \Delta_k \mathds{1}_{\{M \in \mathcal{M}_k\}} + \sum_{k=1}^m \Delta_k \mathds{1}_{\{M \notin \mathcal{M}_k\}} + \sqrt{\frac{5}{8}T \log \frac{8T}{\delta}}  \\
& \leq  D_{aug}\sqrt{\frac{5}{2}T \log \frac{8T}{\delta}} + D_{aug}  SNA \log \frac{8T}{SNA} + (2D_{aug}\\
&\sqrt{14SN  \log(\frac{2AT}{\delta})}  + 2) (\sqrt{2}+1)(\sqrt{SNAT}) +\sqrt{T} + \sqrt{\frac{5}{8}T \log \frac{8T}{\delta}}
\end{split}
\end{equation*}

\normalsize

Further simplifications as in \cite[Appendix C.4]{auer2008near} yield the total regret as :
\footnotesize
\begin{equation*}
 \Delta \leq 34D_{aug}SN  \sqrt{AT\log(T / \delta)}
\end{equation*}
\normalsize
with a probability of $1 - \sum_{T=2}^\infty \frac{\delta}{4T^{5/4}} < 1 - \delta$ by union over all values of $T$.

 \section{PROOF OF THEOREM 2}
 \label{sec:proof_theorem_2}
 
 \subsection{Optimism with concentration inequalities}
\begin{lemma} \label{lemma:pucrlb_M_notin_M_k}
The probability that there exists $k\geq1$ such that  the true AMDP $M$ does not belong to the set of candidate 
AMDP's $\mathcal{M}_k$ denoted by \eqref{eqn:pucrlb_confidence_bound_p}and \eqref{eqn:pucrlb_confidence_bound_r} is at most $\delta/3$
, that is
\begin{equation*}
    \mathbb{P}(\exists k \geq1 s.t.M \notin \mathcal{M}_k )\leq \frac{\delta}{3}
\end{equation*}
\end{lemma}
\begin{proof}
As in \cite[Section 3.2.2]{fruit2019exploration} we bound the probability of the event $E = \cup_{k=1}^{\infty}\{M \notin \mathcal{M}_k\}$. Through out the proof, we use the notation $n_k$ instead of $n_k((s,n),a)$ for brevity. Event $E$ is equivalent to : 
\footnotesize
\begin{eqnarray*}
E & \subseteq \bigcup\limits_{{(s,n),a}} \bigcup\limits_{n_k=0}^{\infty} \{r((s,n),a) \notin \mathcal{B}_r^{k}((s,n),a) \} \\
& \cup \bigcup\limits_{s'}\{p(s'|(s,n),a) \notin \mathcal{B}_p^{k}((s,n),a,s')\} \\
\mathbb{P}(E) & \leq \sum\limits_{{(s,n),a}} \sum\limits_{n_k=0}^{\infty} (\mathbb{P}(r((s,n),a) \notin \mathcal{B}_r^{k}((s,n),a)) \\
& + \sum\limits_{s'}\{\mathbb{P}(p(s'|(s,n),a) \notin \mathcal{B}_p^{k}((s,n),a,s') )
\end{eqnarray*}
\normalsize
where, $\mathcal{B}_r^{k}((s,n),a)$ and $\mathcal{B}_p^{k}((s,n),a,s')$ are as in \eqref{eqn:pucrlb_confidence_bound_p} and \eqref{eqn:pucrlb_confidence_bound_r}.

Let's take a 4-tuple $((s,n),a,s') \in \mathcal{S} \times \mathcal{P} \times \mathcal{A} \times \mathcal{S}$, we define
\footnotesize
\begin{eqnarray*}
\epsilon_{p,k}^{(s,n),a,s'} \hspace{-5mm}& \coloneqq \hat{\sigma}_{p,k}(s'|(s,n),a)\sqrt{\frac{2\log(30S^2NA n_k^2/\delta)}{n_k}} + \frac{3\log(30S^2NA n_k^2/\delta)}{n_k} \\
\epsilon_{r,k}^{(s,n),a} \hspace{-5mm}& \coloneqq \hat{\sigma}_{r,k}((s,n),a)\sqrt{\frac{2\log(30SNA n_k^2/\delta)}{n_k}} + \frac{3\log(30SNA n_k^2/\delta)}{n_k} 
\end{eqnarray*}
\normalsize
Since $\epsilon_{p,k}^{(s,n),a,s'}\leq \beta_{p,k}^{(s,n),a,s'}$ and $\epsilon_{r,k}^{(s,n),a,s'}\leq \beta_{r,k}^{(s,n),a,s'}$, by using Empirical Bernstein Inequality \cite[Theorem 1]{audibert2009exploration}, we can bound the probability of the events as :
\footnotesize
\begin{eqnarray*}
\begin{split}
&  \mathbb{P}\{|\hat{r}_k((s,n),a) - r((s,n),a)| \geq \beta_{r,k}^{(s,n),a} \} \\
& \leq \mathbb{P} \{ |  \hat{r}_k((s,n),a) - r((s,n),a) | \geq \epsilon_{r,k}^{(s,n),a} \} \leq \frac{\delta}{10n_k^2SNA} 
\end{split}
\end{eqnarray*}
\begin{eqnarray*}
\begin{split}
& \mathbb{P} \{ | \hat{p}_{k}(s'|(s,n),a) - p(s'|(s,n),a) | \geq \beta_{p,k}^{(s,n),a} \}\\
& \leq \mathbb{P} \{ |  \hat{p}_{k}(s'|(s,n),a) - p(s'|(s,n),a) | \geq \epsilon_{p,k}^{(s,n),a} \} 
\leq \frac{\delta}{10n_k^2S^2NA} 
\end{split}
\end{eqnarray*}
\normalsize
Thus, 
\footnotesize
\begin{eqnarray*}
\begin{comment}
\mathbb{P}(r((s,n),a) \notin \mathcal{B}_r^n((s,n),a)) \leq \frac{\delta}{10n^2SNA} and \mathbb{P}(p(s'|(s,n),a) \notin \mathcal{B}_p^n((s,n),a,s')) \leq \frac{\delta}{10n^2S^2PA}
\mathbb{P}(\exists T\geq1,\exists k\geq1 s.t. M \notin \mathcal{M}_k)
\end{comment}
\mathbb{P}(E) \leq \sum \limits_{((s,n),a)} (\sum\limits_{n_k=1}^{\infty}(\frac{\delta}{10n_k^2SNA}+\sum\limits_{s'}\frac{\delta}{10n_k^2S^2NA})) = \frac{2\pi^2\delta}{60} \leq \frac{\delta}{3}
\end{eqnarray*}
\end{proof}
\subsection{Splitting into episodes} 

For the stochastic process $ X_t \coloneqq r_t((s_t,n_t),a_t) - r((s_t,n_t), a_t)$, $\{X_t\}_{t\geq1}$ is a Martingale Difference Sequence (MDS) with $\lvert X_t \rvert \leq 1$. Using Azuma's Inequality for MDS \cite[Lemma 10]{auer2008near}, we can write:
\footnotesize
\begin{eqnarray} \label{eqn:pucrlb_MDS} 
\begin{split}
    \mathds{P}(\sum_{t=1}^T X_t \geq \sqrt{4T \log (\frac{4T}{\delta})}) \leq \frac{\delta}{16T^2}
  \end{split}
\end{eqnarray}
\normalsize
Taking a union bound for all possible values of $T\geq1$, with a probability of at least $1- \sum_{T=1}^\infty \frac{\delta}{16T^2} = 1- \frac{\pi^2 \delta}{96}\geq 1- \frac{\delta}{3}$, we obtain:
\footnotesize
\begin{equation*}
    \sum_{t=1}^T X_t \leq \sqrt{4T \log (\frac{4T}{\delta})}
\end{equation*}
\begin{equation*}
    \leftrightarrow \sum_{t=1}^T r_t((s_t,n_t),a_t) \leq \sum_{t=1}^T r((s_t,n_t), a_t) + \sqrt{4T \log (\frac{4T}{\delta})}
\end{equation*}
\normalsize
 Thus we can decompose the total regret as :
\footnotesize
\begin{eqnarray} \label{eqn:pucrlb_regret_decomposition}
\begin{split}
\Delta & = \sum_{t =1}^{T} (\rho^* - r_t((s_t,n_t),a_t))  \leq \sum_{t =1}^{T} (\rho^* -  r((s_t,n_t),a_t)) \\
&+ 2 \sqrt{T \log \frac{4T}{\delta}}  \\
  & \leq \sum_{k=1}^m \Delta_k + 2 \sqrt{T \log \frac{4T}{\delta}}
  \end{split}
\end{eqnarray}
\normalsize
with probability at least $1-\frac{\delta}{3}$, where $m$ represents the total number of episodes and episodic regret $\Delta_k \coloneqq \sum\limits_{(s,n),a} v_k((s,n),a) (\rho^* -  r((s,n),a))$.

\normalsize
\subsection{Episodic Regret}
As in Section~\ref{subsec:M_in_M_k}, with $\epsilon_k = 1/t_k$, the episodic regret can be decomposed as:
\footnotesize
\begin{equation} \label{eqn:pucrlb_episodic_regret_decomposition}
\Delta_{k} \leq \Delta_{k}^{p} + \Delta_{k}^{r}  + 2 \sum_{(s,n),a} \frac{v_k((s,n),a)}{t_k}  
\end{equation}
\normalsize

\subsubsection{Bounding $\Delta_{k}^{p}$}

Following the same arguments as in Section~\ref{sub_sub_Sec:del_p}, the first term of  \eqref{eqn:del_p}, can be bounded similarly to \eqref{eqn:del_p_first_term} as :
 \footnotesize
\begin{equation*} 
{\tau}\sum_{(s,n),a} v_k((s,n),a)( \sum_{s'} w_k(s',n+1)(\tilde{p}_k(s'|(s,n),a) - p_k(s'|(s,n),a))
\end{equation*}
\begin{align*} 
\leq \sum_{(s,n),a} v_k((s,n),a)(  D_{aug}  \lVert  \mathbf{\hat{p}_k}(\cdot|(s,n),a) - \mathbf{p_k}(\cdot|(s,n),a) \rVert_1)
\end{align*}

\begin{align} \label{eqn:pucrlb_del_p_first_term}
\leq \sum_{(s,n),a} v_k((s,n),a)  D_{aug}  \sum_{s'}\beta_{p,k}^{(s,n),a,s'}
\end{align}
\normalsize
 The second term in \eqref{eqn:del_p} after replacing $u_{i_k}$ with $w_k$ can be bounded as : 
 
\begin{equation} \label{eqn:del_p_second_term_pucrlb}
\begin{split}
& \tau \sum_{t=t_k}^{t_{k+1}-1}   ( \sum_{s'} p_k(s'|(s_t,n_t),a) w_k(s',n_{t+1})  - w_k(s_t,n_t))\\
& = \tau \sum_{t=t_k}^{t_{k+1}-1}  ( \sum_{s'} p_k(s'|(s_t,n_t),a) w_k(s',n_{t+1})  \\
& - w_k(s_{t+1},n_{t+1})) \\
& + \tau \sum_{t=t_k}^{t_{k+1}-1}  w_k(s_{t+1},n_{t+1})) - w_k(s_t,n_t)) 
\end{split}
\end{equation}
\normalsize
 The last term in \eqref{eqn:del_p_second_term_pucrlb} is a telescopic sum :
\begin{equation} \label{eq:pucrlb_del_p_third_term}
\begin{split}
    & \tau \sum_{t=t_k}^{t_{k+1}-1}  w_k(s_{t+1},n_{t+1})) - w_k(s_t,n_t)) \\
    & = \tau  ( w_k(s_{t_{k+1}},n_{t_{k+1}})) - w_k(s_{t_{k+1}},n_{t_{k+1}})) ) \\ 
    & \leq \tau span (\mathbf{w}_k) \leq D_{aug}
\end{split}
\end{equation}

Similar to \eqref{eqn:pucrlb_MDS}, for the stochastic process $ X_t \coloneqq  \tau \sum_{s'} p_k(s'|(s_t,n_t),a) w_k(s',n_{t+1}) - \tau w_k(s_{t+1},n_{t+1})$, with $\lvert{X_t}\rvert\leq span (\mathbf{w}_k) \leq D_{aug}^\tau = D_{aug}/\tau$, under event $E^C$, $\forall T \geq 1$, using Azuma's inequality
\begin{equation*}
    \mathds{P}(\sum_{t=1}^T X_t \geq 2 D_{aug}\sqrt{T\frac{4T}{\delta}}) \leq \frac{\delta}{16T^2}
\end{equation*}
 
 Thus, with probability at least $1- \sum_{T=1}^\infty \frac{\delta}{16T^2} \geq 1- \frac{\delta}{3}$
 \begin{equation} \label{eq:pucrlb_del_p_second_term}
 \begin{split}
\tau \sum_{k =1}^{m}  & \sum_{t=t_k}^{t_{k+1}-1}  ( \sum_{s'} p_k(s'|(s_t,n_t),a) w_k(s',n_{t+1})  \\
 & - w_k(s_{t+1},n_{t+1}) \leq 2 D_{aug}\sqrt{T\frac{4T}{\delta}}
 \end{split}
 \end{equation}

Hence, by combining, \eqref{eqn:pucrlb_del_p_first_term}, \eqref{eq:pucrlb_del_p_third_term}, \eqref{eq:pucrlb_del_p_second_term} and substituting $m \leq SNA \log \frac{8T}{SNA}$ as in \cite[Appendix C.2]{auer2008near}, we can write $ \forall T \geq SNA$, under event $E^C$:

\footnotesize
\begin{equation} \label{eqn:pucrlb_del_p_final}
\begin{split}
        \sum_{k =1}^{m} \Delta_{k}^{p} & \leq \sum_{k =1}^{m} \sum_{(s,n),a} v_k((s,n),a)  D_{aug} \sum_{s'} \beta_{p,k}^{(s,n),a,s'} + 2 D_{aug}\sqrt{T\frac{4T}{\delta}} \\
        & + D_{aug} SNA \log \frac{8T}{SNA}
\end{split}
\end{equation}
\normalsize

with a probability of at least $1- \frac{\delta}{3}$.

\subsubsection{Bounding $\Delta_k^r$}
Similar to \ref{sub_sub_Sec:del_r},

\footnotesize
\begin{equation} \label{eqn:pucrlb_del_r}
\Delta_k^r \leq 2\sum_{(s,n),a} v_k((s,n),a) \beta_{r,k}^{(s,n),a} 
\end{equation}

\normalsize
using the confidence bound \eqref{eqn:pucrlb_confidence_bound_r}.

\subsection{Summing over episodes}
 We state a result that would be useful later.
\begin{lemma} \label{lemma:pucrlb_algebraic_results}
    It holds almost surely that $\forall k \geq 1 $ and $\forall ((s,n),a) \in \mathcal{S} \times \mathcal{N} \times \mathcal{A} $:
    \begin{equation*}
    \begin{split}
               & \sum_{k =1}^{m}  \frac{v_k((s,n),a)}{\sqrt{n_k((s,n),a)}} \leq 3 \sqrt{n_{m+1}((s,n),a)} \\
        & \sum_{k =1}^{m}  \frac{v_k((s,n),a)}{n_k((s,n),a)} \leq 2 + 2 \log{n_{m+1}((s,n),a)} 
    \end{split}
    \end{equation*}
\end{lemma}
\begin{proof}
    Refer \cite[Lemma 3.6]{fruit2019exploration}.
\end{proof}

Under event $E^C$, combining the results of Lemma~\ref{lemma:pucrlb_M_notin_M_k} ,\eqref{eqn:pucrlb_episodic_regret_decomposition}, \eqref{eqn:pucrlb_del_p_final} and \eqref{eqn:pucrlb_del_r}, we can bound the total sum of episodic regret $\forall T \geq SNA$ as :

\begin{equation} \label{eqn:pucrlb_episodic_regret_decomposition_final}
\begin{split}
\sum_{k =1}^{m} \Delta_{k} & \leq \underbrace{ D_{aug}  \sum_{k =1}^{m} 
 \sum_{(s,n),a} v_k((s,n),a)   \sum_{s'} \beta_{p,k}^{(s,n),a,s'}}_{\doteq \Delta_4} \\
 & + \underbrace{\sum_{k =1}^{m} 2\sum_{(s,n),a} v_k((s,n),a) \beta_{r,k}^{(s,n),a}}_{\doteq \Delta_5} \\ 
 & + \underbrace{2 \sum_{k =1}^{m} \sum_{(s,n),a} \frac{v_k((s,n),a)}{t_k}}_{\doteq \Delta_6} 
 + 2 D_{aug}\sqrt{T\frac{4T}{\delta}} \\ 
 & + D_{aug} SNA \log \frac{8T}{SNA}
 \end{split}
\end{equation}
with a probability of at least $1- \frac{2\delta}{3}$.

\subsubsection{Bounding $\Delta_4$}
Using \eqref{eqn:pucrlb_beta_p} and ${n_k((s,n),a)} \leq T$, $\Delta_4$ can be bounded as :
\begin{equation*}
    \begin{split}
        \sum_{k =1}^{m} & \sum_{(s,n),a} v_k((s,n),a) \sum_{s'}  \beta_{p,k}^{(s,n),a,s'} \leq 2 \sqrt{\log(\frac{6SNAT}{\delta})} \\
        &\sum_{k =1}^{m} \sum_{(s,n),a} \frac{v_k((s,n),a)}{\sqrt{n_k((s,n),a)}} \\
        & \sum_{s'}\sqrt{\hat{p}_k(s'|(s,n),a) (1-\hat{p}_k(s'|(s,n),a))} \\
        &+ 6S\log(\frac{6SNAT}{\delta})\sum_{k =1}^{m} \sum_{(s,n),a} \frac{v_k((s,n),a)}{n_k((s,n),a)}
    \end{split}
\end{equation*}

\begin{lemma} \label{lemma:pucrlb_del_p_variance_bound}
It holds almost surely that for all $k \geq 1$ and for all $((s,n),a,s') \in \mathcal{S} \times \mathcal{P} \times \mathcal{A} \times \mathcal{S}$:
\begin{equation*}
    \sum_{s'}\sqrt{\hat{p}_k(s'|(s,n),a) (1-\hat{p}_k(s'|(s,n),a))} \leq \sqrt{\Gamma^{\mathcal{S}}((s,n),a)}
\end{equation*}
\end{lemma}
\begin{proof}
    Refer \cite[Appendix A.2]{fruit2019exploration}.
\end{proof}

Thus using Lemma \ref{lemma:pucrlb_algebraic_results} and \ref{lemma:pucrlb_del_p_variance_bound} and the results in \cite[Section 3.5.6]{fruit2019exploration} :

\begin{equation} \label{eqn:pucrlb_del_p_final_2}
    \begin{split}
        \sum_{k =1}^{m} & \sum_{(s,n),a} v_k((s,n),a)  \sum_{s'} \beta_{p,k}^{(s,n),a,s'} \leq \\
         & 6 \sqrt{(\sum_{(s,n),a} \Gamma^{\mathcal{S}}((s,n),a))T\log(\frac{6SNAT}{\delta})} \\
        &+ 12S^2NA\log(\frac{6SNAT}{\delta})(1+\log{T})
    \end{split}
\end{equation}

\subsubsection{Bounding $\Delta_5$}
Using \eqref{eqn:pucrlb_beta_r}, $\Delta_5$ can be bounded as :
\begin{equation*}
    \begin{split}
        \sum_{k =1}^{m} & \sum_{(s,n),a} v_k((s,n),a)   \beta_{r,k}^{(s,n),a} \leq \sum_{k =1}^{m} \sum_{(s,n),a} \\
        & 2 \sqrt{\hat{\sigma}_{r,n_k}^2((s,n),a) \log(\frac{6SNA n_k((s,n),a)}{\delta})} \frac{v_k((s,n),a)}{\sqrt{n_k((s,n),a)}} \\
        & + 6ln(\frac{6SNA n_k((s,n),a)}{\delta}) \frac{v_k((s,n),a)}{n_k((s,n),a)}
    \end{split}
\end{equation*}
By using Lemma \ref{lemma:pucrlb_algebraic_results}, ${n_k((s,n),a)} \leq T$ and the fact that $\hat{\sigma}_{r,n_k}^2((s,n),a) \leq 1$ as reward is in [0,1]:
\begin{equation} \label{eqn:pucrlb_del_r_final}
    \begin{split}
        \sum_{k =1}^{m} & \sum_{(s,n),a} v_k((s,n),a)   \beta_{r,k}^{(s,n),a} \leq 6 \sqrt{ SNAT\log(\frac{6SNAT}{\delta})} \\
        & + 12 SNA \log(\frac{6SNAT)}{\delta})(1+\log{T})
    \end{split}
\end{equation}

\subsubsection{Bounding $\Delta_6$}
Since $t_k \geq n_k((s,n),a) \forall ((s,n),a)$ and using Lemma \ref{lemma:pucrlb_algebraic_results}, $\Delta_6$ can be bounded as :
\begin{equation} \label{eqn:pucrlb_third_term_final}
    \begin{split}
        \sum_{k =1}^{m} \sum_{(s,n),a} \frac{v_k((s,n),a)}{t_k} & \leq \sum_{k =1}^{m} \sum_{(s,n),a} \frac{v_k((s,n),a)}{n_k((s,n),a)} \\
        & \leq SNA (2+2\log{T})
    \end{split}
\end{equation}

Combining \eqref{eqn:pucrlb_regret_decomposition}, \eqref{eqn:pucrlb_episodic_regret_decomposition_final},\eqref{eqn:pucrlb_del_p_final_2}, \eqref{eqn:pucrlb_del_r_final}, \eqref{eqn:pucrlb_third_term_final}, and taking a union bound, we can bound the total regret $\forall T \geq SNA$ and under event $E^C$ as :

\footnotesize
\begin{eqnarray} 
\begin{split}
\Delta & \leq 6 \sqrt{(\sum_{(s,n),a} \Gamma^{\mathcal{S}}((s,n),a))T\log(\frac{6SNAT}{\delta})} \\
        &+ 12S^2NA\log(\frac{6SNAT}{\delta})(1+\log{T}) + 6 \sqrt{ SNAT\log(\frac{6SNAT}{\delta})} \\
        & + 12 SNA \log(\frac{6SNAT)}{\delta})(1+\log{T}) + 4 SNA (1+\log{T}) \\ & + 2 D_{aug}\sqrt{T\frac{4T}{\delta}}  + D_{aug} SNA\log_2 \frac{8T}{SNA} + 2 \sqrt{T \log \frac{4T}{\delta}}
  \end{split}
\end{eqnarray}
\normalsize

  with probability at least $1-\delta$.
For some positive constant $\beta$, this is equivalent to :

\footnotesize
\begin{eqnarray*} 
\begin{split}
\Delta & \leq \beta D_{aug} \{\sqrt{(\sum_{(s,n),a} \Gamma^{\mathcal{S}}((s,n),a))T\log(\frac{T}{\delta})} \\& + S^2NA\log(\frac{T}{\delta}) \log(T)\}
  \end{split}
\end{eqnarray*}
\normalsize

Since   $\Gamma^{\mathcal{S}}((s,n),a)) \leq S$ , we obtain $\sum_{(s,n),a} \Gamma^{\mathcal{S}}((s,n),a)) \leq S^2NA.$

  This gives the final bound as:
  \footnotesize
\begin{eqnarray*} 
\begin{split}
\Delta & \leq \beta D_{aug} [S\sqrt{NAT\log(\frac{T}{\delta})} + S^2NA\log(\frac{T}{\delta}) \log(T) ]
\end{split}
\end{eqnarray*}
\end{document}